\documentclass[10pt,journal,letterpaper,compsoc,twocolumn]{IEEEtran}

\usepackage[nocompress]{cite}
\usepackage{graphicx}

\usepackage[cmex10]{amsmath}
\usepackage[linesnumbered,boxed,commentsnumbered]{algorithm2e}
\usepackage{array}
\usepackage[tight,normalsize,sf,SF]{subfigure}
\usepackage{amsthm}                      
\usepackage{multirow}                    
\usepackage{cite}
\usepackage{graphicx}
\usepackage{psfrag}
\usepackage{subfigure}
\usepackage{url}
\usepackage{stfloats}
\usepackage{amssymb,amsmath}
\usepackage{amsfonts}
\usepackage{array}

\begin{document}

\title{Towards Big Topic Modeling}

\author{Jian-Feng~Yan,~\IEEEmembership{Member,~IEEE,}
        Jia~Zeng,~\IEEEmembership{Senior Member,~IEEE,
        Zhi-Qiang~Liu,
        and~Yang~Gao~}%
\IEEEcompsocitemizethanks{\IEEEcompsocthanksitem Jian-Feng Yan, Jia~Zeng and Yang~Gao are with the School of Computer Science and Technology,
Soochow University, Suzhou 215006, China.
Jia Zeng is the corresponding author.
E-mail: j.zeng@ieee.org.\protect\\
}%
\IEEEcompsocitemizethanks{\IEEEcompsocthanksitem Zhi-Qiang Liu is with the School of Creative Media,
City University of Hong Kong, Tat Chee Ave. 83, Kowloon Tong, Hong Kong, P.R. China.\protect\\}%

\thanks{}}%


\IEEEcompsoctitleabstractindextext{%

\begin{abstract}
To solve the big topic modeling problem,
we need to reduce both time and space complexities of batch latent Dirichlet allocation (LDA) algorithms.
Although parallel LDA algorithms on the multi-processor architecture have low time and space complexities,
their communication costs among processors often scale linearly with the vocabulary size and the number of topics,
leading to a serious scalability problem.
To reduce the communication complexity among processors for a better scalability,
we propose a novel communication-efficient parallel topic modeling architecture based on power law,
which consumes orders of magnitude less communication time when the number of topics is large.
We combine the proposed communication-efficient parallel architecture with the online belief propagation (OBP) algorithm referred to as POBP for big topic modeling tasks.
Extensive empirical results confirm that POBP has the following advantages to solve the big topic modeling problem:
1) high accuracy,
2) communication-efficient,
3) fast speed,
and 4) constant memory usage
when compared with recent state-of-the-art parallel LDA algorithms on the multi-processor architecture.
\end{abstract}

\begin{keywords}
Big topic modeling, latent Dirichlet allocation, communication complexity, multi-processor architecture, online belief propagation, power law.
\end{keywords}}

\maketitle

\IEEEdisplaynotcompsoctitleabstractindextext

\section{Introduction}

Probabilistic topic modeling~\cite{Steyvers:07,Blei:12}
provides a powerful method for data analysis in machine learning and applied statistics.
In this paper,
we study one of the most successful topic modeling algorithms,
latent Dirichlet allocation (LDA)~\cite{Blei:03},
which has been wildly used in many fields such as text mining, computer vision and computational biology.
Big topic modeling algorithms have attracted intensive research interests
because big data have become increasingly common in recent years such as billions of tweets, images and videos on the web.

However,
it is still a big challenge to reduce both time and space complexities of traditional batch LDA algorithms
such as variational Bayes (VB)~\cite{Blei:03},
collapsed Gibbs sampling (GS)~\cite{Griffiths:04},
and belief propagation (BP)~\cite{Zeng:11} for big topic modeling tasks.
For example,
if we use the batch BP~\cite{Zeng:11} to extract $10,000$ topics from the PUBMED data set containing $8.2$ million documents~\cite{Porteous:08},
the memory to store all documents and LDA parameters takes around $36$ TB,
and the time consumption for $200$ iterations is around $3$ months on a single processor.
Therefore,
both time and space costs are unaffordable in many real-world applications.
Recent big topic modeling solutions fall into three categories:
1) fast batch LDA algorithms,
2) online LDA algorithms,
and 3) parallel LDA algorithms.

Fast batch LDA algorithms observe the fact that the probability mass of the topic distribution
is concentrated only on a small set of the topics when the number of topics is very large.
This sparseness property facilitates fast Gibbs sampling (FGS)~\cite{Porteous:08} and sparse Gibbs sampling (SGS)~\cite{sgs} algorithms.
The basic idea is to sample a topic by checking the topics with high concentrated probability mass first.
Generally,
FGS and SGS run around $8\sim20$ times faster than traditional GS~\cite{Griffiths:04} when the number of topics is very large.
Active belief propagation (ABP)~\cite{zeng2012new} is a sublinear BP algorithm~\cite{Zeng:11} for topic modeling.
At each iteration,
it scans only a subset of topics and documents for a fast convergence speed.
In practice,
ABP is around $10\sim20$ times faster than SGS or FGS to reach convergence with a higher topic modeling accuracy.
Despite of the fast speed on large data sets,
anchor word recovery-based topic modeling algorithms~\cite{Arora:13} scale nonlinearly with the vocabulary size and the number of topics.
Although a significant speedup has been achieved,
these fast batch LDA algorithms still require a large memory space to store both data and LDA parameters.

Unlike fast batch solutions,
online LDA algorithms require only a constant memory space by treating both data and LDA parameters as streams composed of several small mini-batches.
After sequentially loading each mini-batch into memory for computation until convergence,
we free each mini-batch from memory after one look.
In practice,
we need to confirm that online algorithms can converge to the local optimum point of LDA's objective function.
Within the stochastic optimization framework~\cite{Bottou:book},
online variational Bayes (OVB)~\cite{Hoffman:10} and online belief propagation (OBP)~\cite{Zeng:14} have been proved to fulfill this goal.
Generally,
online algorithms are faster than their batch counterparts by a factor of $2$ to $5$ due to fast local gradient descents.
However,
online algorithms rarely use the powerful parallel architectures to further scale their performances because of high communication costs or serious race conditions~\cite{Yan:09,Ahmed:12}.

Parallel LDA algorithms use the widely available parallel architecture to speed up topic modeling process.
Currently,
there are two types of parallel architectures: multi-processor~\cite{Newman:09} and multi-core~\cite{Yan:09},
where the difference lies in the way to use memory.
In the multi-processor architecture (MPA),
all processes have separate memory spaces and communicate to synchronize LDA parameters at the end of each iteration.
In the multi-core architecture (MCA),
all threads share the same memory space so that race condition is serious.
There are three important questions remaining to be addressed in recent parallel LDA algorithms:
\begin{enumerate}
\item
Accuracy: Can parallel LDA algorithms produce the same results as those of batch counterparts on a single processor?
\item
Communication cost: How to reduce the communication cost in MPA?
\item
Race condition: How to alleviate the race condition in MCA?
\end{enumerate}
Almost all parallel GS (PGS) algorithms~\cite{Wang:09,Ahmed:12,Liu:11,Newman:09,Porteous:08,Yan:09,Smola:10}
can yield only an approximate result with that of batch GS~\cite{Griffiths:04},
while the parallel VB (PVB)~\cite{Zhai:12} is able to produce exactly the same result with that of batch VB~\cite{Blei:03}.
To alleviate race conditions on the GPU MCA,
a streaming approach is proposed to partition data into several non-conflict data streams in memory~\cite{Yan:09}.
But this partition process may introduce the loading imbalance problem for a low parallel efficiency.
As far as MPA is concerned,
the reduction of communication cost still remains an unsolved problem
since the communication cost is often too big to be masked by computation time in web-scale applications.
The experimental results confirm that the communication cost may exceed the computation cost
to become the primitive cost of big topic modeling~\cite{Wang:09,Liu:11}.
Therefore,
in this paper we focus on reducing the communication complexity in MPA for big topic modeling tasks.
It is not difficult to combine MPA and MCA for a better parallel architecture to solve big topic modeling problems.

To achieve the communication-efficient goal,
we propose a novel MPA based on the power law~\cite{Newman:05},
which has a few orders of magnitude less communication cost when compared with the current state-of-the-art
parallel LDA algorithms~\cite{Newman:09,Porteous:08,Yao:09,Ahmed:12,Zhai:12}.
Besides,
we combine this parallel architecture with the current state-of-the-art online LDA algorithm OBP~\cite{Zeng:14}
referred to as POBP for big topic modeling tasks with the following advantages:
\begin{enumerate}
\item Convergence to the local optimum of the LDA's objective function;
\item Communication-efficient;
\item Fast speed;
\item Constant memory usage.
\end{enumerate}
In experiments,
our POBP runs $5\sim100$ times faster,
uses constant memory space,
consumes around $5\% \sim 20\%$ communication time,
but achieves $20\%\sim65\%$ higher topic modeling accuracy than current state-of-the-art parallel LDA algorithms.
Therefore,
we anticipate that the proposed communication-efficient MPA scheme can be generalized to other parallel machine learning algorithms.

The remainder of this paper is organized as follows.
Section~\ref{s2} reviews 1) online belief propagation (OBP) algorithm~\cite{Zeng:14} and
2) the current MPA scheme~\cite{Newman:09} for big topic modeling.
Section~\ref{s3} presents our solution POBP and introduces how to use power law to significantly reduce the communication complexity in MPA.
Section~\ref{s4} compares the proposed POBP with several state-of-the-art parallel LDA algorithms.
Finally,
Section~\ref{s5} makes conclusions and envisions further work.

\section{Related Work} \label{s2}

\begin{table}[t]
\centering
\caption{Notations.}
\begin{tabular}{|l|l|} \hline
$1 \le d \le D$                      &Document index                 \\ \hline
$1 \le w \le W$                      &Word index in vocabulary       \\ \hline
$1 \le k \le K$                      &Topic index                    \\ \hline
$1 \le m \le M$                      &Mini-batch index               \\ \hline
$1 \le n \le N$                      &Processor index                \\ \hline
$\mathbf{x}_{W \times D} = \{x_{w,d}\}$     &Document-word matrix           \\ \hline
$\mathbf{z}_{W \times D} = \{z^k_{w,d}\}$   &Topic labels for words         \\ \hline
$\boldsymbol{\theta}_{K \times D}$          &Document-topic distribution    \\ \hline
$\boldsymbol{\phi}_{K \times W}$            &Topic-word distribution        \\ \hline
$\alpha,\beta$                       &Dirichlet hyperparameters      \\ \hline
\end{tabular} \label{notation}
\end{table}

We briefly review OBP~\cite{Zeng:14} and MPA~\cite{Newman:09} for big topic modeling.
We show that a simple combination of OBP and MPA will cause unaffordable communication costs for a bad scalability performance.
Table~\ref{notation} summarizes important notations in this paper.

LDA allocates a set of thematic topic labels,
$\mathbf{z} = \{z^k_{w,d}\}$,
to explain non-zero elements in the document-word co-occurrence matrix $\mathbf{x}_{W \times D} = \{x_{w,d}\}$,
where $1 \le w \le W$ denotes the word index in the vocabulary,
$1 \le d \le D$ denotes the document index in the corpus,
and $1 \le k \le K$ denotes the topic index.
Usually,
the number of topics $K$ is provided by users.
The nonzero element $x_{w,d} \ne 0$ denotes the number of word counts at the index $\{w,d\}$.
For each word token $x_{w,d,i} = \{0,1\}, x_{w,d} = \sum_i x_{w,d,i}$,
there is a topic label $z^k_{w,d,i} = \{0,1\}, \sum_{k=1}^K z^k_{w,d,i} = 1, 1 \le i \le x_{w,d}$.
We define the soft topic label for the word index $\{w,d\}$ by $z_{w,d}^k = \sum_{i=1}^{x_{w,d}} z^k_{w,d,i}x_{w,d,i}/x_{w,d}$,
which is an average topic labeling configuration over all word tokens at index $\{w,d\}$.
The objective of LDA is to maximize the joint probability $p(\mathbf{x},\boldsymbol{\theta},\boldsymbol{\phi}|\alpha,\beta)$,
where $\boldsymbol{\theta}_{K \times D}$ and $\boldsymbol{\phi}_{K \times W}$
are two non-negative matrices of multinomial parameters for document-topic and topic-word distributions,
satisfying $\sum_k \theta_d(k) = 1$ and $\sum_w \phi_w(k) = 1$.
Both multinomial matrices are generated by two Dirichlet distributions with hyperparameters $\alpha$ and $\beta$.
For simplicity,
we consider the smoothed LDA with fixed symmetric hyperparameters~\cite{Griffiths:04}.

\subsection{OBP}

Online belief propagation (OBP)~\cite{Zeng:14} combines active belief propagation (ABP)~\cite{Zeng:13}
with stochastic gradient descent framework~\cite{Bottou:book}.
It partitions the document-word matrix $\mathbf{x}_{W\times D}$ into mini-batches $x^m_{w,d}, 1 \le d \le D_m, 1 \le m \le M$.
After loading the $m$th mini-batch into memory,
OBP infers the posterior probability called message $\sum_k\mu^m_{w,d}(k) = 1$,
$\mu^m_{w,d}(k) = p(z^{k,m}_{w,d,i} = 1|x^m_{w,d,i}=1,\boldsymbol{\theta},\boldsymbol{\phi};\alpha,\beta)$,
\begin{align} \label{message}
\mu^m_{w,d}(k) \propto \frac{[\hat{\theta}^m_{-w,d}(k) + \alpha] \times [\hat{\phi}^m_{w,-d}(k) + \beta]}{\hat{\phi}^m_{-(w,d)}(k) + W\beta},
\end{align}
where $\hat{\theta}$ and $\hat{\phi}$ are the {\em sufficient statistics} for the online LDA model,
\begin{gather}
\label{theta}
\hat{\theta}^m_{-w,d}(k) = \sum_{-w} x_{w,d}^m\mu_{w,d}^m(k), \\
\label{phi}
\hat{\phi}^m_{w,-d}(k) = \hat{\phi}^{m-1}_w(k) + \sum_{-d} x^m_{w,d}\mu^m_{w,d}(k),
\end{gather}
where $-w$ and $-d$ denote all word indices except $w$ and all document indices except $d$,
and $-(w,d)$ denotes all indices except $\{w,d\}$.
The multinomial parameters of document-topic and topic-word distributions
$\boldsymbol{\theta}$ and $\boldsymbol{\phi}$ can be obtained by normalizing
sufficient statistics $\boldsymbol{\hat{\theta}}$ and $\boldsymbol{\hat{\phi}}$.
Each mini-batch is swept for several iterations $T_m$ until the convergence condition is reached.
Then,
OBP frees from memory the $m$th mini-batch, the local $\mu^m_{w,d}(k)$ and $\hat{\theta}^m_{-w,d}(k)$.
The global topic-word distribution $\phi^m_w(k)$ in memory will be re-used by the next mini-batch.
When the size of $\phi^m_w(k)$ is very large,
we may also store the entire matrix in hard disk and load the partial matrix in memory for computation~\cite{Zeng:14}.

OBP is an ideal choice for big stream topic modeling on the single-processor platform because of several advantages.
First,
OBP guarantees convergence to the stationary point of LDA's likelihood function
within the online expectation-maximization (EM) framework~\cite{Neal:98,Liang:09,Olivier:09}.
Second,
OBP is memory-efficient by using disk as the storage extension.
Its space complexity in memory is proportional to the mini-batch size $D_m$ and the number of topics $K$.
Finally,
OBP is built upon time-efficient ABP algorithm~\cite{Zeng:13},
whose time complexity is insensitive to the number of topics $K$ and the number of documents in each mini-batch $D_m$.
However,
the communication complexity is intractable if we directly parallelize OBP in MPA for big topic modeling tasks,
which will be explained in detail in the next subsection.

\subsection{MPA}

\begin{figure*}[t]
\centering
\includegraphics[width=0.5\linewidth]{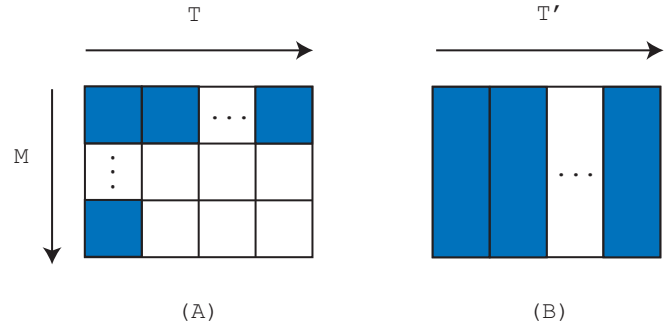}
\caption{
A comparison of communication costs between parallel (A) online and (B) batch LDA algorithms.
Each blue box denotes a communication operation.
In (A),
the communication rate depends on the number of iterations over all mini-batches $MT$,
while in (B),
the communication rate depends on the number of iterations $T'$.
}
\label{mpa}
\end{figure*}

The MPA scheme has been widely used in many parallel batch LDA algorithms~\cite{Wang:09,Liu:11,Newman:09,Porteous:08,Yao:09,Zhai:12}.
Here,
we extend this scheme to parallelize online LDA algorithms.
The MPA~\cite{Newman:09} distributes each mini-batch $\mathbf{x}_{W \times D_m}$ documents over $1 \le n \le N$ processors.
The processor $n$ gets approximately $D_{m,n} = D_m/N$ documents.
The local $\boldsymbol{\hat{\theta}}^{m,n}_{K \times D_{m,n}}$ can be also distributed into $N$ processors,
but the global $\boldsymbol{\hat{\phi}}^{m,n}_{K \times W}$ have to be shared by $N$ processors
since each distributed mini-batch $\mathbf{x}_{W \times D_{m,n}}$ may still cover the entire vocabulary words.
After sweeping each mini-batch $\mathbf{x}_{W \times D_{m,n}}$ at the end of each iteration $1 \le t \le T_m$,
the $N$ processors have to communicate and synchronize the global matrix
$\hat{\phi}^{m,.,t}_{K \times W}$ from $N$ local matrices $\hat{\phi}^{m,n,t}_{K \times W}$ by
\begin{align} \label{syn1}
\hat{\phi}^{m,.,t}_w(k) = &\hat{\phi}^{m,.,t-1}_w(k) + \notag \\
&\sum_{n=1}^N[\hat{\phi}^{m,n,t}_w(k) - \hat{\phi}^{m,.,t-1}_w(k)].
\end{align}
Then,
the synchronized matrix $\hat{\phi}^{m,.,t}_w(k)$ is transferred to each processor
to replace $\hat{\phi}^{m,n,t}_w(k)$ for the next mini-batch.
Thus,
the communication complexity is
\begin{align} \label{cc}
\text{Communication complexity} \propto NMTKW,
\end{align}
where $N$ is the number of processors,
$M$ the number of mini-batches,
$K$ the number of topics,
$W$ the vocabulary size,
and $T = \sum_{m=1}^M T_m/M$ the average number of iterations to reach convergence for each mini-batch.
For example,
suppose that we use $1000$ processors to learn $K=2000$ topics with $T = 100$ from the PUBMED data set~\cite{Porteous:08}
having $W = 141,043$ and $M = 500$ mini-batches.
The total communication cost reaches around $100$ PB ($10^{15}$ bytes) according to~\eqref{cc}.
Meanwhile,
the time complexity of OBP reduces linearly with the number of processors $N$.
So,
the communication cost will be greater than the computation cost when $N \rightarrow \infty$.
In this situation,
adding more processors will not reduce the entire topic modeling time, leading to serious scalability issues.
The major reason why MPA still works in previous parallel batch LDA algorithms~\cite{Wang:09,Liu:11,Newman:09,Porteous:08,Yao:09}
is that the communication cost depends only on the number of batch iterations $T'$ rather than the number of iterations over mini-batches $MT$,
where practically $T' \ll MT$.
If $T' = 500$,
the parallel batch LDA algorithms require only $1$PB communication cost in the above example,
which is significantly smaller than that of parallel online LDA algorithms.
For some real-world big data streams,
the number of mini-batches may reach infinity~\cite{Zhai:13},
i.e.,
$M \rightarrow \infty$.
Thus, communication cost of parallel online LDA algorithms may become so huge as to seriously damage parallel efficiency.

Fig.~\ref{mpa} compares the communication costs between parallel batch and online LDA algorithms.
Parallel batch LDA algorithms communicate and synchronize $\boldsymbol{\hat{\phi}}_{K \times W}$ at the end of each batch iteration,
while parallel online algorithms do it at the end of each mini-batch iteration.
Generally,
the number of batch iterations $T'$ is significantly smaller than the number of mini-batch iterations $MT$.
Thus,
the higher communication rate leads to the larger communication cost in parallel online LDA algorithms.
Therefore,
it is nontrivial to reduce the communication complexity~\eqref{cc}
for parallel online LDA algorithms~\cite{Yao:09,Hoffman:10,Wahabzada:11,Mimno:12,Zeng:14}
in order to achieve a better scalability performance.
Moreover,
not all parallel batch LDA algorithms based on MPA have been proved to converge to the local optimum of the LDA's objective function.
Typical examples include those GS-based parallel algorithms~\cite{Wang:09,Liu:11,Newman:09,Porteous:08,Yao:09} in MPA framework.

\section{POBP} \label{s3}

In this paper,
we propose a communication-efficient MPA and explain this scheme using power law~\cite{Newman:05}.
Combining with OBP,
we propose the parallel OBP (POBP) to solve the big topic modeling problem.
We show that POBP has low time, space and communication complexities,
and can converge to the local optimum of the LDA's objective function within the online EM framework~\cite{Neal:98,Liang:09,Olivier:09}.

\subsection{Communication-Efficient MPA} \label{s3.1}

\begin{figure*}[t]
\centering
\includegraphics[width=0.6\linewidth]{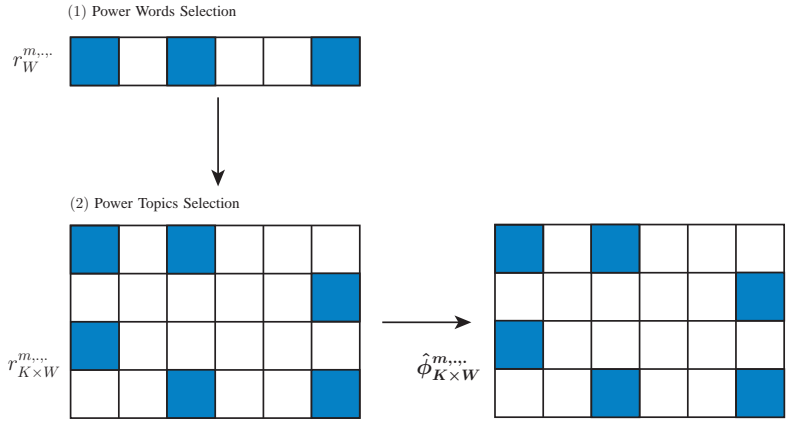}
\caption{
The two-step power words and topics selection process
for a global matrix $\boldsymbol{\hat{\phi}}_{K \times W}$ with $K=4$ and $W=6$,
where $\lambda_K = \lambda_W = 0.5$.
The blue boxes denote the selected power words and topics.
In the first step,
we select power words by sorting the synchronized residual vector $r^{m,t}_w$.
In the second step,
for each selected power word we select further power topics by sorting the synchronized residual matrix $r^{m,t}_w(k)$ in $K$ dimensions.
}
\label{selection}
\end{figure*}

\begin{figure*}[t]
\centering
\includegraphics[width=0.8\linewidth]{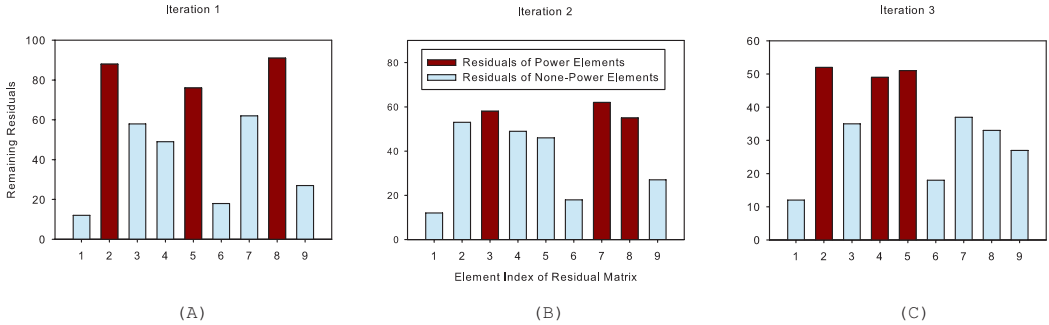}
\caption{
A dynamic scheduling example of a residual matrix $r_{3 \times 3}$ with $3$ words and $3$ topics,
where the $9$ elements in $r_{3 \times 3}$ are shown in one dimension.
(A) In the first iteration $t=1$, the elements $\{2,5,8\}$ are chosen as power elements.
(B) In the second iteration $t=2$, the elements $\{3,7,8\}$ are chosen as power elements because residuals of $\{2,5\}$ become relatively smaller.
(C) In the third iteration, the elements $\{2,4,5\}$ are selected as the power elements because residuals of $\{3,7,8\}$ become relatively smaller.
}
\label{residualevolution}
\end{figure*}

From~\eqref{cc},
there are two straight-forward solutions to reduce the communication cost.
The first is to reduce the average communication rate $T$.
For example,
we may communicate and synchronize the global matrix at every two mini-batch iterations to reduce around half communication cost.
This heuristic solution has been widely used in MPA~\cite{Newman:09} but with two problems:
1) the lower communication rate may cause the lower topic modeling accuracy;
and 2) the overall communication rate depends also on the number of mini-batches $M$,
which is often constrained by each processor's memory space.
Therefore,
we investigate the second solution to communicate and synchronize only the subset of global matrix at each mini-batch iteration,
i.e.,
reduce the size $KW$ in~\eqref{cc}.
To our best knowledge,
there are very few investigations in related work following this research line.
We will further explain why selecting the subset of global matrix
dynamically does not influence the topic modeling accuracy very much based on power law.

We propose a two-step strategy to select the subset of global matrix at each iteration in a dynamic manner.
First,
we select a subset of vocabulary words with size $\lambda_WW$ referred to as the {\em power words}.
For each power word,
we select a subset of topics with size $\lambda_KK$ referred to as the {\em power topics}.
In this way,
we reduce the communication complexity~\eqref{cc} from $KW$ to $\lambda_K\lambda_WKW$ as follows,
\begin{align} \label{cc1}
\text{Communication complexity} \propto \lambda_K\lambda_WNMTKW,
\end{align}
where the ratios $0 < \lambda_K \ll 1$ and $0< \lambda_W \ll 1$.
Obviously,
Eq.~\eqref{cc1} shows a sublinear complexity of~\eqref{cc}.
The remaining question is how to select both power words and topics.

Our selection criterion is inspired by the residual belief propagation (RBP)~\cite{Elidan:06,Zeng:12a}.
At each processor $n$,
we define the residual between message vectors~\eqref{message} at two successive iterations $t$ and $t-1$,
\begin{gather} \label{residual}
r^{m,n,t}_{w,d}(k) = x^{m,n}_{w,d}|\mu^{m,n,t}_{w,d}(k) - \mu^{m,n,t-1}_{w,d}(k)|, \\
r^{m,n,t}_w(k) = \sum_d r^{m,n,t}_{w,d}(k).
\end{gather}
We then communicate and synchronize the residual matrix $r^{m,n,t}_w(k)$ across $N$ processors similar to~\eqref{syn1},
\begin{align} \label{residual1}
r^{m,.,t}_w(k) = &r^{m,.,t-1}_w(k) + \notag \\
&\sum_{n=1}^N[r^{m,n,t}_w(k) - r^{m,.,t-1}_w(k)].
\end{align}
From~\eqref{residual1},
we further obtain the synchronized residual vector of vocabulary words,
\begin{align} \label{residual2}
r^{m,.,t}_w = \sum_k r^{m,.,t}_w(k).
\end{align}
Finally,
we sort vector~\eqref{residual2} in the descending order,
and select the power words with $\lambda_WW$ largest residuals.
For each power word,
we sort matrix~\eqref{residual1} in the $K$ dimension,
and select $\lambda_KK$ power topics for each word with largest residuals.

Fig.~\ref{selection} shows an example of the two-step selection method for the global matrix $\boldsymbol{\hat{\phi}}_{4 \times 6}$.
We set the selection ratios as $\lambda_K = \lambda_W = 0.5$.
In the first step,
we select three power words with largest residuals in the vector $r^{m,.,t}_w$ denoted by the blue boxes.
In the second step,
for each selected power word,
we select two power topics with largest residuals in the matrix $r^{m,.,t}_w(k)$ in the $K$ dimension.

This two-step selection process follows the dynamical scheduling scheme.
For $m$th mini-batch at the first iteration $t=1$,
we need to communicate and synchronize the entire matrices $\boldsymbol{\hat{\phi}}^{m,.,1}_{K \times W}$ and $r^{m,.,1}_{K \times W}$.
When $2 \le t \le T_m$,
we communicate and synchronize only the partial matrices
$\boldsymbol{\hat{\phi}}^{m,.,2 \le t \le T_m}_{\lambda_KK \times \lambda_WW}$ and $r^{m,.,2 \le t \le T_m}_{\lambda_KK \times \lambda_WW}$,
while we keep the remaining elements untouched.
Residuals~\eqref{residual} of power words and topics are getting smaller and smaller in the message passing process according to Eq.~\eqref{message}.
Therefore,
the power words and topics in the previous iteration may be no longer power ones due to their relatively smaller residuals in the next iteration.
In this way,
all vocabulary words and topics have the chances to be selected as power ones before convergence.
When all elements in residual matrix reach zeros, i.e.,
$r^{m,.,t}_w(k) \rightarrow 0$,
the message passing process reaches the convergence state.

For a better understanding of the dynamic scheduling process,
Fig.~\ref{residualevolution} shows an example $r^{t=1,2,3}_{3 \times 3}$ at different iterations,
where the nine elements are shown in one dimension for simplicity.
Fig.~\ref{residualevolution}A shows that in the first iteration,
the elements $\{2,5,8\}$ are selected as the power elements to pass messages such that the residuals for the three elements decrease while other residuals remain unchanged.
Fig.~\ref{residualevolution}B shows that elements $\{3,7,8\}$ are selected as the power elements in the second iteration
because the elements $\{2,5\}$ get relatively smaller residuals.
However,
they could be power elements again in next iterations when their residuals become relatively higher than those of other elements.
Fig.~\ref{residualevolution}C shows that the elements $\{2,4,5\}$ are chosen as power elements in the third iteration.
Therefore,
we can guarantee that no information gets lost since all elements have chance to become power elements to pass messages,
which ensures the topic modeling accuracy of the algorithm.

\subsection{The POBP Algorithm}

\begin{figure*}[t]
\centering
\includegraphics[width=1.0\linewidth]{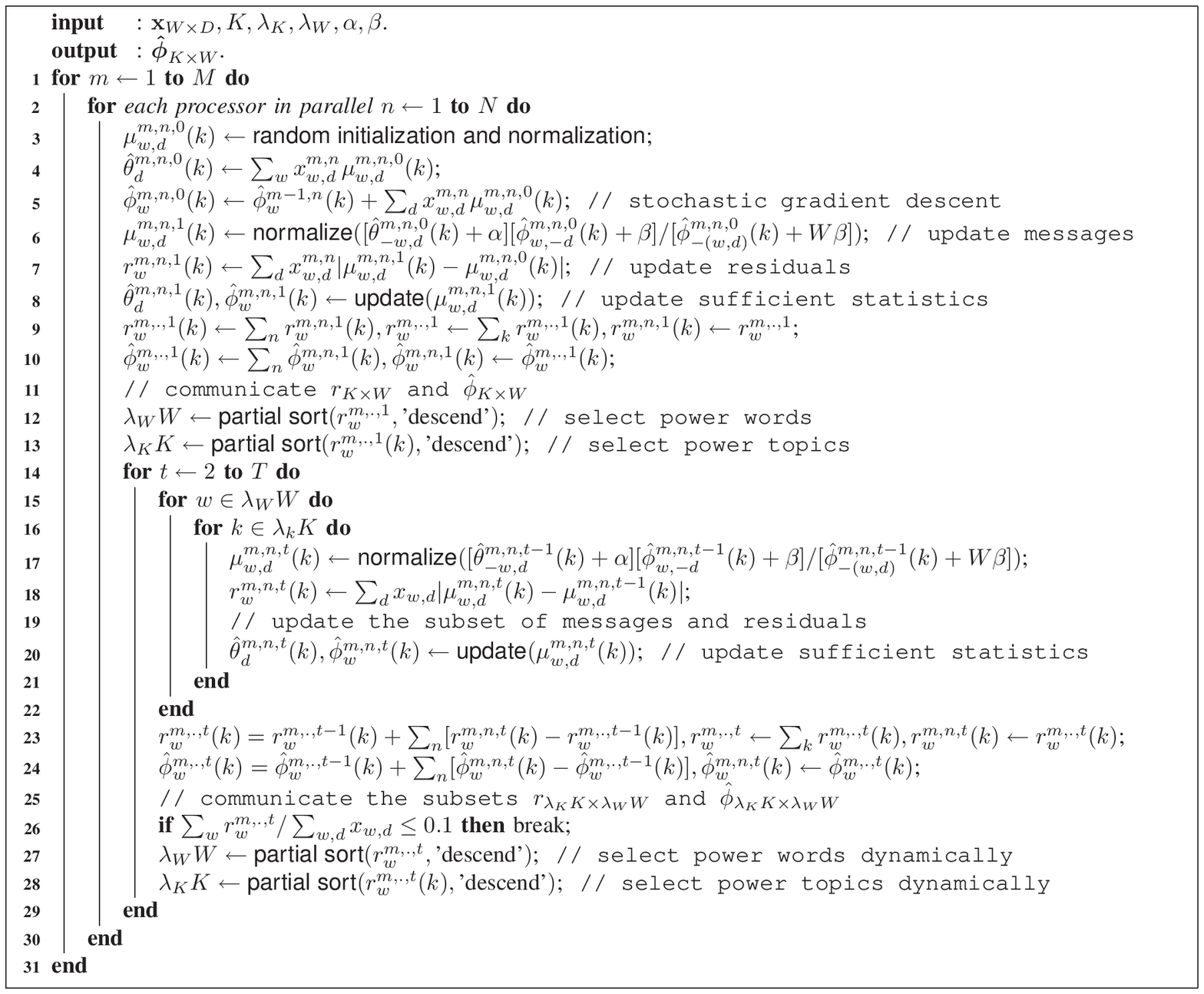}
\caption{
The POBP algorithm for LDA.
}
\label{pobpcode}
\end{figure*}

Although we focus on developing parallel online belief propagation (POBP) algorithm for big topic modeling tasks in this subsection,
the proposed communication-efficient MPA can be applied to both parallel batch and online LDA algorithms.
Fig.~\ref{pobpcode} summarizes the proposed POBP algorithm.
We distribute each incoming mini-batch $x^{m,n}_{w,d}$ into $N$ processors in parallel (line $2$).
At the first iteration $t=1$,
we random initialize and normalize messages $\mu^{m,n,0}_{w,d}$ (line $3$),
which are used to update sufficient statistics $\hat{\theta}^{m,n,0}_d(k)$ and $\hat{\phi}^{m,n,0}_w(k)$
using Eqs.~\eqref{theta} and~\eqref{phi} (lines $4$ and $5$).
Note that we use the stochastic gradient descent~\cite{Robbins:51,Bottou:book} to update~\eqref{phi} in line $5$,
where the initial $\hat{\phi}^{m=0}$ is set as the zero matrix.
Then,
we update both messages $\mu^{m,n,1}_{w,d}(k)$ and residuals $r^{m,n,1}_w(k)$ using Eqs.~\eqref{message} and~\eqref{residual} (lines $6$ and $7$).
The messages are in turn used to update sufficient statistics $\hat{\theta}^{m,n,1}_d(k)$ and $\hat{\phi}^{m,n,1}_w(k)$ (line $8$).
At the end of the first iteration,
all processors communicate and synchronize two global matrices $\hat{\phi}^{m,.,1}_w(k)$ and $r^{m,.,1}_w(k)$,
and transfer the global matrices back to each processor (lines $9$ and $10$).
Using two-step selection method,
we select the power words and topics from the global residual matrix (lines $12$ and $13$).
We use the partial sort to find the power words and topics with top largest $\lambda_WW$ and $\lambda_KK$.
The computation cost of partial sort algorithm is significantly lower than
quick sort since we do not need the complete sorting.
Also,
we use the parallel implementations of partial sort algorithm to further speed up the selection process.
The time complexity of partial sort is at most $W \log W$ and $K \log K$,
where $W$ is the vocabulary size and $K$ is the number of topics.

In the following iterations $2 \le t \le T$,
we update only the subsets of messages $\mu^{m,n,t}_{w,d}(k)$ and $r^{m,n,t}_w(k)$ residuals based on the selected power words and topics (lines $17$ and $18$),
and communicate only the subsets of matrices $\hat{\phi}^{m,.,t}_w(k)$ and $r^{m,.,t}_w(k)$ (lines $23$ and $24$).
In the dynamical scheduling process,
we select the power words and topics based on the synchronized residual matrix $r^{m,.,t}_w(k)$ (lines $27$ and $28$).
If the average of the residual matrix is blow a threshold (line $26$),
we terminate all processors and load the next mini-batch $x^{m+1,n}_{w,d}$
after freeing memory except for the global topic-word matrix $\hat{\phi}^{m,.,.}_w(k)$.
POBP terminates until all $M$ mini-batches have been processed (line $1$).
When $M \rightarrow \infty$,
POBP can be viewed as a life-long or never-ending topic modeling algorithm.
The output is the global sufficient statistics $\boldsymbol{\hat{\phi}}_{K \times W}$,
which can be normalized to obtain the topic-word multinomial parameter matrix $\boldsymbol{\phi}_{K \times W}$.
If $N = 1$,
POBP reduces to the OBP~\cite{Zeng:14} algorithm on a single processor.
If $M = 1$,
POBP reduces to the parallel batch BP algorithm on $N$ processors~\cite{Zeng:12b}.

\subsubsection{Convergence Analysis} \label{s3.2.1}

The objective of LDA is to maximize the joint probability
$p(\mathbf{x},\boldsymbol{\theta},\boldsymbol{\phi}|\alpha,\beta)$~\cite{Blei:03,Asuncion:09,Zeng:11}.
According to the MAP inference~\cite{Freitas:01,Chien:08,Asuncion:09},
this objective can be achieved by the iterative EM algorithm~\cite{Zeng:14},
where the E-step has almost the same message update equation with~\eqref{message},
and the M-step resembles Eqs.~\eqref{theta} and~\eqref{phi}.
OBP uses the stochastic gradient descent method~\cite{Robbins:51,Bottou:book} to update the topic-word matrix,
\begin{align} \label{gradient}
\hat{\phi}^m_w(k) = \hat{\phi}^{m-1}_w(k) + \frac{1}{m-1}\Delta \hat{\phi}^m_w(k),
\end{align}
where $\Delta \hat{\phi}^m_w(k) = \sum_d x^m_{w,d}\mu^m_{w,d}(k)$ in Eq.~\eqref{phi} is the gradient generated by the current mini-batch.
Eq.~\eqref{gradient} has a learning rate $1/(m-1)$ because
$\hat{\phi}^{m-1}_w(k)$ accumulates sufficient statistics of previous $m-1$ mini-batches,
and $\Delta \hat{\phi}^m_w(k)$ accumulates only sufficient statistics of the current mini-batch.
The parameter estimation is invariant to the scaling of sufficient statistics~\eqref{phi}.
Since this learning rate satisfies two conditions,
\begin{gather}
\label{obprate1}
\sum_{m = 2}^{\infty} \frac{1}{m-1} = \infty, \\
\label{obprate2}
\sum_{m = 2}^{\infty} \frac{1}{(m - 1)^2} < \infty,
\end{gather}
the online stochastic approximation~\cite{Robbins:51} shows that sufficient statistics $\hat{\phi}^m_w(k)$ will converge to a stationary point,
and the gradient $\Delta \hat{\phi}^m_w(k)$ will converge to zero when $m \rightarrow \infty$.
Using~\eqref{gradient},
OBP can incrementally improve $\boldsymbol{\hat{\phi}}^m$ to maximize the log-likelihood $\ell(\cdot)$ of the joint probability of LDA
within the online EM framework~\cite{Neal:98,Liang:09,Olivier:09},
\begin{align} \label{convergeobp}
\ell(\boldsymbol{\hat{\phi}}^{m+1}) \ge \ell(\boldsymbol{\hat{\phi}}^{m}).
\end{align}
More detailed proof of~\eqref{convergeobp} can be referred to~\cite{Zeng:14}.
In this sense,
when $m \rightarrow \infty$,
OBP can converge to the local optimum of the LDA's log-likelihood function.

Similarly,
we show that POBP in Fig.~\ref{pobpcode} can also achieve this goal on $N$ processors.
As far as the $m$th mini-batch is concerned,
the global $\hat{\phi}^{m-1}_w(k)$ of the previous mini-batch remains unchanged for $N$ processors (line $5$).
Indeed,
all processors update just the local gradient $\Delta \hat{\phi}^{m,n,t}_w(k)$ from the current mini-batch $x^{m,n}_{w,d}$,
and communicate this local gradient according to~\eqref{syn1} as follows,
\begin{align} \label{syn2}
\Delta \hat{\phi}^{m,.,t}_w(k) = &\Delta \hat{\phi}^{m,.,t-1}_w(k) + \notag \\
&\sum_{n=1}^N[\Delta \hat{\phi}^{m,n,t}_w(k) - \Delta \hat{\phi}^{m,.,t-1}_w(k)],
\end{align}
where the synchronized gradient is almost the same with~\eqref{gradient}.
Also,
the learning rate is still $1/(m-1)$,
which guarantees the convergence of POBP.
If we do not communicate at each iteration,
Eq.~\eqref{syn2} produces the inaccurate local gradient~\eqref{gradient},
and thus leads to a slow convergence speed.
However,
from~\eqref{gradient} and~\eqref{syn2},
lowering the communication rate does not change the convergence property of POBP,
but reduces its convergence speed.
The proposed POBP communicates more frequently than its offline counterparts as shown in Fig.~\ref{mpa},
which ensures its superiority over offline algorithms in terms of convergence performance.

\subsubsection{Complexity and Scalability} \label{s3.2.2}

\begin{table*}[t]
\centering
\caption{Comparison of complexities.}
\begin{tabular}{|c|c|c|c|} \hline
Algorithms &Computation cost &Memory cost &Communication cost
\\ \hline \hline
POBP  &$\eta\lambda_K\lambda_WKWDT/N$ &$K(\eta WD + D)/MN + 2KW$  &$\lambda_K\lambda_WKWMNT$ \\
OBP\cite{Zeng:14} &$\eta\lambda_K\lambda_WKWDT$ &$K(\eta WD + D)/M + 2KW$ &$-$ \\
PGS~\cite{Newman:09} &$\eta'KWDT'/N$ &$(K \times D + \eta' WD)/N + KW$ &$NKWT'$  \\
\hline
\end{tabular}
\label{complexities}
\end{table*}

Table~\ref{complexities} compares the complexities of POBP with those of OBP~\cite{Zeng:14} and PGS~\cite{Newman:09} algorithms.
For simplicity,
we assume that the number of non-zero element in $\mathbf{x}_{W \times D}$ is $\eta WD$,
where $\eta$ is a very small constant value depending on the data sets because $\mathbf{x}_{W \times D}$ is very sparse.
Similarly,
we assume that the total number of word tokens in $\mathbf{x}_{W \times D}$ is $\eta' WD = \sum_{w,d}x_{w,d}$,
where $\eta'$ is also a constant value depending on data sets.
Generally,
$\eta \ll \eta'$ for most data sets.
Suppose that the overall computation cost is $A$,
and the communication cost for each processor is $B$,
and thus the overall cost of $N$ processors can be simplified as
\begin{align} \label{overallcost}
\text{Overall cost} = \frac{A}{N} + BN,
\end{align}
where
\begin{align} \label{numberofprocessors}
N^* = \sqrt{\frac{A}{B}},
\end{align}
minimizes the overall cost~\eqref{overallcost} to $2\sqrt{AB}$.
From~\eqref{numberofprocessors},
we see that it is the ratio between computation and communication costs that determines the scalability,
i.e.,
the best number of processors for the minimum overall cost.
Note that in practice the communication cost per processor $B$ is a variable that depends also on the bandwidth limitation between processors.
When $N$ increases,
$B$ will also increase nonlinearly due to complex communication operations over limited bandwidth.
Although Eq.~\eqref{overallcost} is a simplified estimation of relationship between computation and communication costs,
it provides clues for estimation of the optimal number of processors in practice.
For simplicity,
we use~\eqref{overallcost} and~\eqref{numberofprocessors} in the following analysis,
where we use the size of communicated and synchronized matrices of each processor in Table~\ref{complexities} to approximate $B$.

For each mini-batch at the first iteration ($t=1$),
POBP requires to scan the entire mini-batch and communicate two complete matrices
$\boldsymbol{\hat{\phi}}_{K \times W}$ and $r_{K \times W}$.
In the following iterations,
POBP scans only the subset of mini-batch,
and communicate the subsets of matrices $\boldsymbol{\hat{\phi}}_{K \times W}$ and $r_{K \times W}$.
Since the number of iterations for convergence is often very large (for example, $T \approx 200$),
the total computation and communication costs are dominated by the rest iterations ($2 \le t \le T$).
So,
we approximate the overall computation and communication costs (without considering the small partial sorting costs) shown in Table~\ref{complexities}.
The real-world costs are proportional to these values.
According to~\eqref{overallcost} and~\eqref{numberofprocessors},
the best number of processors is
\begin{align} \label{pobpprocessors}
N^* \propto \sqrt{\frac{\eta D}{M}} = \sqrt{\eta D_m},
\end{align}
and the minimal overall cost is
\begin{align} \label{minicost}
\text{POBP's minimum cost} \propto 2\lambda_K\lambda_WKWT\sqrt{\eta DM}.
\end{align}
This analysis is consistent with our intuition that the best number of processors in POBP scales linearly with the mini-batch size $D_m$.
When $M = 1$,
POBP reduces to the parallel batch BP algorithm with the minimum overall cost when the best number of processors reaches the maximum.

However,
the memory cost of each processor becomes very high shown in Table~\ref{complexities}
because we have to store the local message matrix $\boldsymbol{\mu}_{K \times \eta WD}$,
the document-topic matrix $\boldsymbol{\hat{\theta}}_{K \times D/M}$,
the global topic-word matrix $\boldsymbol{\hat{\phi}}_{K \times W}$ and the residual matrix $r_{K \times W}$.
Therefore,
POBP provides a flexible solution by setting the number of mini-batches $M$ for big topic modeling tasks.
When each processor has enough memory space,
we can set the smaller number of mini-batches $M$ and use more processors for the fast speed.
When there is no enough memory for each processor,
we can set larger number of mini-batches $M$ and use less processors for a relatively slow speed.
Note that the minimum overall cost of POBP scales with the square root $\sqrt{DM}$,
which is often significantly lower than that of the OBP (e.g., OBP scales linearly with the number of documents $D$)
on a single processor shown in Table~\ref{complexities}.
Moreover,
POBP uses less memory of each processor than OBP.
In this sense,
POBP is more suitable than OBP for big topic modeling tasks in real-world applications.

Table~\ref{complexities} also shows the complexities of PGS algorithm~\cite{Newman:09},
which is one of the widely-used big topic modeling solutions introduced in Section~\ref{s2}.
Its computation scales linearly with the number word tokens in $\mathbf{x}_{W \times D}$.
According to~\eqref{overallcost} and~\eqref{numberofprocessors},
the best number of processors is $\sqrt{\eta' D}$
and the minimum overall cost is $2KWT'\sqrt{\eta'D}$.
Obviously,
POBP often has the lower minimum cost~\eqref{minicost} than that of PGS.
Moreover,
POBP consumes less memory than PGS,
so that it is more suitable for big topic modeling tasks.
Indeed,
if $\lambda_W = 0.1$ and $\lambda_K = 50/K$ (See experiments in subsection~\ref{s4.1}),
POBP's minimum cost is insensitive to the number of topics $K$ and the vocabulary size.
This is a good property since big data sets often contain a big number of topics and vocabulary words~\cite{Zhai:13}.
Although parallel FGS (PFGS)~\cite{Porteous:08} and SGS (PSGS)~\cite{Yao:09} are also insensitive to the number of topics $K$,
they still consume more memory space than POBP.
Also,
lowering the computation cost instead of communication cost will make the scalability worse as shown in Eq.~\eqref{numberofprocessors},
i.e.,
the best number of processors $N^*$ will become smaller.

\subsection{Power Law Explanation}

\begin{figure}[t]
\centering
\includegraphics[width=0.8\linewidth]{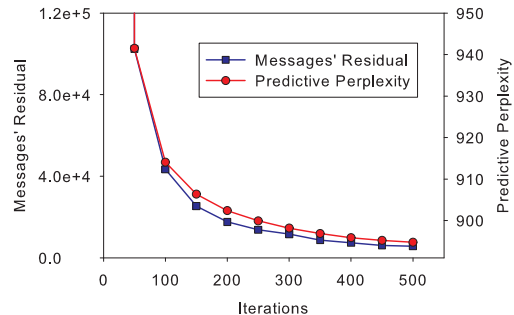}
\caption{
The residual (blue curve) and predictive perplexity (red curve) as a function of iterations on ENRON.
The predictive perplexity goes down with the residual,
which indicates the convergence.
}
\label{residualperplexity}
\end{figure}

\begin{figure*}[t]
\centering
\includegraphics[width=1\linewidth]{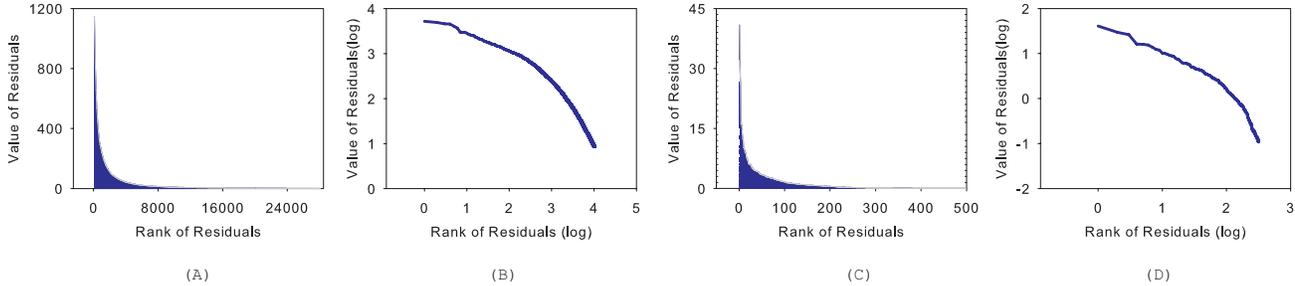}
\caption{
The message value as a function of the message rank when $K=500$ at the $10$th iteration on ENRON.
(A) Linear plot for message rank for vocabulary words.
(B) Log-log plot for message rank for vocabulary words.
(C) Linear plot for message rank for topics.
(D) Log-log plot for message rank for topics.
}
\label{powerlawresidual}
\end{figure*}

Power law,
also known as the long-tail principle or the $80/20$ rule~\cite{Sanders:92},
refers to the fact that a major proportion of effects come from a small fraction of the causes for many events.
We show that the selected power words and topics based on residuals~\eqref{residual1} and~\eqref{residual2} follow power law.
In this paper,
we take the ENRON data set~\cite{Porteous:08} as an example to show the appearance of power law.
We set the number of topics as $500$ and select the messages at the $10$th iteration.

First,
we show that the residual~\eqref{residual} can evaluate the convergence of topic modeling process,
where the predictive perplexity~\eqref{perplexity} has been widely used as the convergence condition of LDA algorithms~\cite{Blei:03,Asuncion:09,Zeng:11}.
Fig.~\ref{residualperplexity} compares the predictive perplexity of LDA and the average residual over all words.
We see that the two curves have almost the same trend reflecting the convergence state.
This is why we use the average residual as the convergence condition in the POBP algorithm in Fig.~\ref{pobpcode} (line $26$).
Intuitively,
if residuals become zeros,
the message values do not change so that the parameters are fixed at the local optimum.
In this sense,
we can speed up convergence by minimizing the larger residuals first and then the smaller residuals.
This is the first motivation of our two-step selection method in communication-efficient MPA in subsection~\ref{s3.1}.

Second,
we show that the distribution of residuals approximately follows power law at each mini-batch iteration.
A simple way to identify power-law behavior in either natural or man-made systems
is to draw a histogram with both axis plotted on logarithmic scales called log-log plot~\cite{Sanders:92}.
If the log-log plot approximates a straight line,
we consider that power law applies.
We sort the residuals in descending order.
We draw the linear plot for $r_w$ in Fig.~\ref{selection} with the x-axis for residual ranks and y-axis for residual values.
Fig.~\ref{powerlawresidual}A indicates that a small fraction of words take a vary large proportion of residuals.
Fig.~\ref{powerlawresidual}B shows that the corresponding log-log plot approximately follows power law.
This phenomenon confirms that only a small subset of vocabulary words contribute almost all residual values.
More specifically,
the top $10\%$ words account for $79\%$ of the total residual value,
while the top $20\%$ words account for almost $90\%$ of the total residual value.
Therefore,
it is efficient to minimize residuals of those power words fist to speed up the convergence.
Fig.~\ref{powerlawresidual}C shows the linear plot for $r_w(k)$ in Fig.~\ref{selection},
and Fig.~\ref{powerlawresidual}D shows the corresponding log-log plot.
Both confirm that the residual distribution of power topics approximately follows power law.
Therefore,
we only need to do computation and communication for power words and topics,
which will be updated through the dynamical scheduling at each iteration in Fig.~\ref{residualevolution}.

\section{Experiments} \label{s4}

\begin{table*}[t]
\centering
\caption{Summarization of four data sets.}
\begin{tabular}{|c|c|c|c|c|c|c|c|c|} \hline
Data sets  &$D$          &$W$                        &$N_{\text{token}}$        &$NNZ$            &Size (M) \\
\hline \hline
ENRON      &$39,861$     &$6,536$                    &$6,412,172$               &$2,374,385$      &$28.34$ \\
NYTIMES    &$300,000$    &$7,871$                    &$99,542,125$              &$44,379,275$     &$568.88$ \\
WIKIPEDIA  &$4,360,095$  &$5,363$                    &$665,375,061$             &$154,934,308$    &$1983.77$\\
PUBMED     &$8,200,000$  &$6,902$                    &$737,869,083$             &$222,399,377$    &$3043.04$ \\
\hline
\end{tabular}
\label{datasets}
\end{table*}

We compare the proposed POBP with parallel FGS (PFGS)~\cite{Porteous:08},
parallel SGS (PSGS)~\cite{Yao:09},
Yahoo LDA (YLDA)~\cite{Ahmed:12},
and parallel variational Bayes (PVB)~\cite{Zhai:12}.
All these benchmark algorithms have open source codes.
For a fair comparison,
we re-write their source codes in C++ language~\cite{Zeng:12}.
Also,
we use the integer type to store LDA parameters in the GS-based algorithms,
while we use the single-precision floating-point format to represent LDA parameters in both PVB and POBP algorithms.
Such an implementation difference is caused by the sampling process in the GS-based algorithms~\cite{Griffiths:04}.

We run the above algorithms on a cluster with up to $1024$ processors ($1.9$GHz CPU, $2$GB memory) to perform the experiments.
All the processors communicate through a high-speed Infiniband with $20$GB per second bandwidth.
Following~\cite{Porteous:08},
we use the fixed hyper-parameters $\alpha = 2/K$ and $\beta = 0.01$ for all algorithms to guarantee a fair comparison.
To reach the convergence state,
we run PFGS, PSGS, YLDA and PVB using $500$ iterations~\cite{Newman:09}.
For POBP,
we set $NNZ \approx 45,000$ in each-mini batch since OBP's performance is insensitive to the mini-batch size~\cite{Zeng:14}.
Also,
this mini-batch size can be easily fit into $2$GB memory of each processor.
We evenly distribute $D$ documents to $N$ processors to avoid load imbalance.

We use four publicly available data sets: ENRON, NYTIMES, PUBMED~\cite{Porteous:08}
and WIKIPEDIA,\footnote{\url{http://en.wikipedia.org}}
where ENRON is a relatively smaller data set so that we use it for parameters tuning.
The other three data sets are relatively bigger with up to $8$ million documents and we use them for web-scale experiments.
We follow~\cite{Hoffman:10} and remove the words out of a fixed truncated vocabulary to get a shorter vocabulary
because some vocabulary words occur rarely and contribute little to topic modeling.
While the vocabulary size $W$ has been greatly reduced,
most of the word tokens $N_{\text{token}}$ and none-zero-elements $NNZ$ are still reserved.
For example,
though we reduce the vocabulary size of PUBMED from $141,043$ to $6,902$ with a ratio of $4.89\%$,
we reduce the number of word tokens from about $7$ millon to $3$ millon with a ratio over $40\%$.
As a result,
we can fit the word-topic distribution $\boldsymbol{\phi}_{K \times W}$ in $2$GB memory of each processor when $K$ is large.
Table~\ref{datasets} summarizes the statistics of data sets,
where $D$ denotes the number of documents,
$W$ the vocabulary size,
$N_{\text{token}}$ the number of word tokens,
$NNZ$ the number of non-zero elements,
and ``Size (M)" size of data sets in MByte.

We use the predictive perplexity ($\mathcal{P}$)~\cite{Asuncion:09,Zeng:11} to measure accuracy of different parallel LDA algorithms.
To calculate the predictive perplexity,
we randomly partition each document into $80\%$ and $20\%$ subsets.
Fixing the word-topic distribution $\boldsymbol{\phi}_{K \times W}$,
we estimate $\boldsymbol{\theta}_{K \times D}$ on the $80\%$ subset by the training algorithms from the same random initialization after $500$ iterations,
and then calculate the predictive perplexity on the rest $20\%$ subset,
\begin{align} \label{test}
\mathcal{P} = \exp\Bigg\{-\frac{\sum_{w,d}
x_{w,d}^{20\%}\log\big[\sum_{k}\theta_d(k)\phi_w(k)\big]}
{\sum_{w,d} x_{w,d}^{20\%}}\Bigg\},
\end{align}
where $x_{w,d}^{20\%}$ denotes word counts in the the $20\%$ subset.
The lower predictive perplexity represents a higher accuracy.

\subsection{Ratios $\lambda_W$ and $\lambda_K$} \label{s4.1}

\begin{figure*}[t]
\centering
\includegraphics[width=0.9\linewidth]{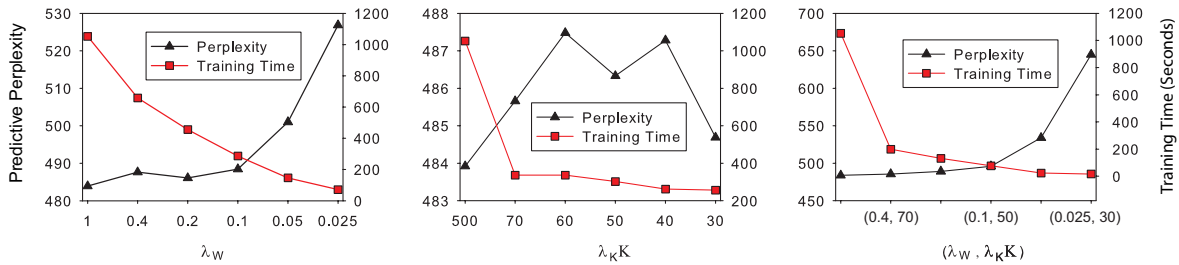}
\caption{
Predictive perplexity and training time as a function of $\lambda_K$ and $\lambda_W$ on ENRON data set.
We fix $K=500$ and use $12$ processors.
The left axis denotes the predictive perplexity and the right axis denotes the training time in seconds.
(A) Fixing $\lambda_K = 1$, we test different $\lambda_W = \{0.025, 0.05, 0.1, 0.2, 0.4,1\}$.
(B) Fixing $\lambda_W = 1$, we test different $\lambda_KK = \{30, 40, 50, 60, 70,500\}$.
(C) We test some combinations of $\lambda_W$ and $\lambda_KK$.
We see that when $\lambda_W = 0.1$ and $\lambda_KK = 50$
POBP can achieve a significant speedup while achieving a good accuracy.
}
\label{set}
\end{figure*}

POBP introduces two parameters $\lambda_W$ and $\lambda_K$ to control the ratio of power words and topics at each iteration.
The parameter $\lambda_K$ determines the ratio of power topics evolved at each iteration.
The smaller $\lambda_K$ will lead to less computation and communication cost.
However,
this may also result in a lower topic modeling accuracy.
In practice,
each word may not be allocated to many topics,
and thus $\lambda_KK$ is often a fixed value.
To study the effect of different $\lambda_KK$,
we evaluate a range of $\lambda_KK$ values on the ENRON data set when $K = 500$.

Fig.~\ref{set}A shows the predictive perplexity and training time as a function of $\lambda_W$ by fixing $\lambda_K = 1$,
where $\lambda_W = 1$ denotes that all the vocabulary words are scanned at each iteration.
We decrease the value of $\lambda_W$ from $0.4$ to $0.025$ in an exponential manner.
While the training time decreases with the decrease of $\lambda_W$,
the predictive perplexity also increases indicating a degraded performance.
However,
when $\lambda_W \ge 0.1$,
the increase of perplexity is so small that can be neglected.
This result confirms that a subset of power words at each iteration contributes to almost all topic modeling performance.
Also,
we see that a small value of $\lambda_W$ may lead to an increase of perplexity.
For example,
when $\lambda_W =0.025$,
the predictive perplexity increases around $8\%$ to $526.8$.

Fig.~\ref{set}B shows the predictive perplexity and training time as a function of $\lambda_KK$ by fixing $\lambda_W = 1$,
where $\lambda_KK=500$ means that all the topics are scanned at each iteration.
We change $\lambda_KK$ from $30$ to $70$ in a step of $10$.
The results show that the predictive perplexity increases slightly and the training time decreases steadily with the decrease of $\lambda_KK$.
Fig.~\ref{set}B also confirms that a subset of power topics plays an important role in topic modeling.
Finally,
we combine different values of $\lambda_W$ and $\lambda_KK$.
Fig.~\ref{set}C shows that $\{\lambda_W = 0.1, \lambda_K K = 50\}$
can achieve a reasonable speedup while keeping a high accuracy
(e.g., the predictive perplexity change is within $15$).
We also use this setting in subsection~\ref{s3.2.2} for complexity and scalability analysis.

\subsection{Accuracy}

\begin{figure*}[t]
\centering
\includegraphics[width=0.8\linewidth]{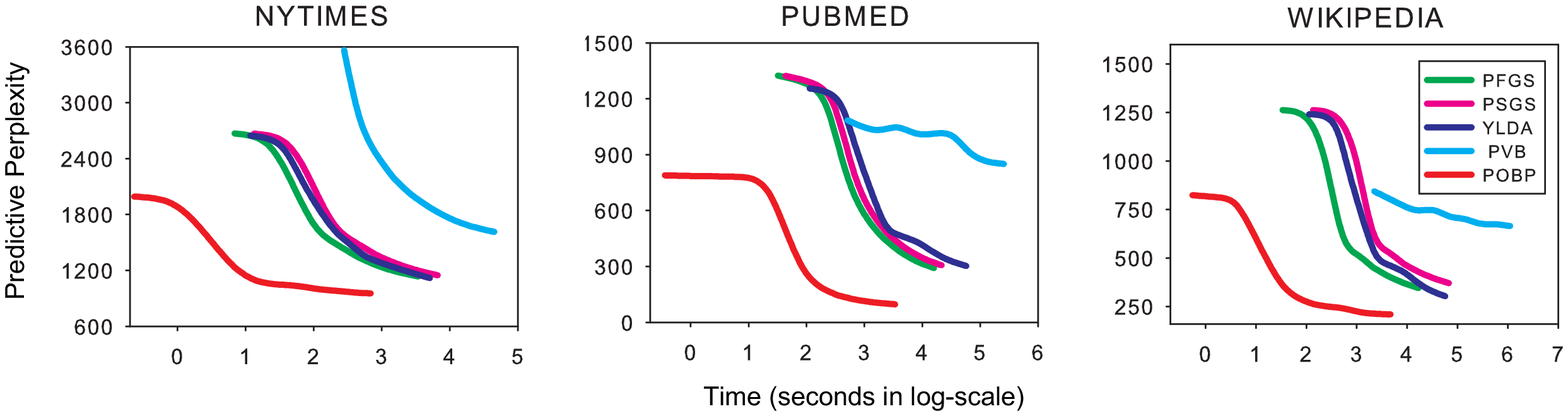}
\caption{
Predictive perplexity as a function of training time (in second log-scale) on NYTIMES, PUBMED
and WIKIPEDIA data sets using $256$ processors when $K = 2000$.
}
\label{convergence}
\end{figure*}

\begin{figure*}[t]
\centering
\includegraphics[width=0.8\linewidth]{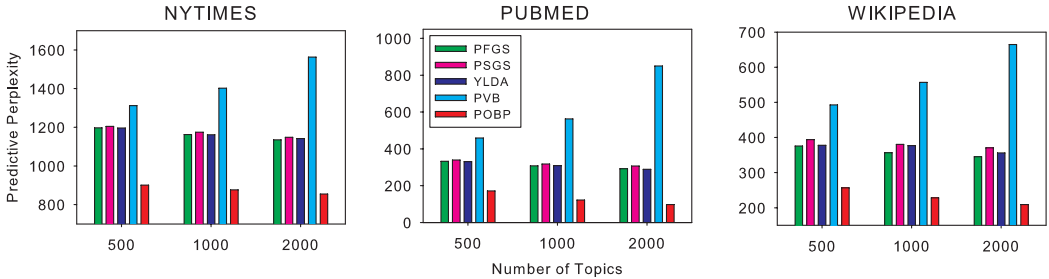}
\caption{
Comparison of predictive perplexity for all algorithms on NYTIMES, PUBMED and WIKIPEDIA data sets using $256$ processors,
where the number of topics $K \in \{500, 1000, 2000\}$.
}
\label{perplexity}
\end{figure*}

\begin{table}[t]
\centering
\caption{Perplexity gap between POBP and PFGS.}
\begin{tabular}{|c|c|c|c|c|} \hline
$K$                     &NYTIMES          &WIKIPEDIA       &PUBMED       \\ \hline \hline
$500$                   &$24.41\%$        &$31.64\%$       &$48.54\%$    \\
$1000$                  &$24.57\%$        &$36.07\%$       &$60.46\%$    \\
$2000$                  &$24.69\%$        &$39.51\%$       &$66.68\%$    \\ \hline
\end{tabular}
\label{gap}
\end{table}

Fig.~\ref{convergence} shows the predictive perplexity as a function of training time (in second log-scale)
on NYTIMES, PUBMED and WIKIPEDIA using $256$ processors when $K = 2000$.
We see that POBP converges fastest among all the algorithms,
around $10$ to $100$ times faster than GS-based algorithms and $50$ to $400$ times faster than PVB.
This result is consistent with our convergence analysis in subsection~\ref{s3.2.1}.
Also,
POBP always reaches the lowest predictive perplexity,
indicating its good convergence property.
Fig.~\ref{perplexity} also shows that POBP yields the lowest predictive perplexity on all data sets
given different number of topics on $256$ processors.
The GS-based algorithms such as PFGS, PSGS and YLDA have slightly higher perplexity while PVB produces the highest perplexity.
These results are consistent with observations in previous work~\cite{Asuncion:09,Zeng:11,Zeng:13}.
We see that the predictive perplexity of PVB increases with the number of topics partly due to the overfitting phenomenon.

Table~\ref{gap} compares the perplexity gap between POBP and PFGS calculated by
\begin{align} \label{gap_cal}
\text{gap} = {\frac{\mathcal{P}_{\text{PFGS}} - \mathcal{P}_{\text{POBP}}} {\mathcal{P}_{\text{PFGS}}}} \times 100\%,
\end{align}
where $\mathcal{P}$ is the predictive perplexity~\eqref{test}.
When $K = 500$,
the gap is about $24.41\%$ on relatively smaller data set NYTIMES
but the gap increases to $31.64\%$ and $48.54\%$ on larger data sets WIKIPEDIA and PUBMED, respectively.
Besides,
the gap increases for all data sets when $K$ increases from $500$ to $2000$.
Such an excellent predictive performance
makes POBP a very competitive topic modeling algorithm on real-world big data streams.

\subsection{Communication Time}

\begin{figure*}[t]
\centering
\includegraphics[width=0.8\linewidth]{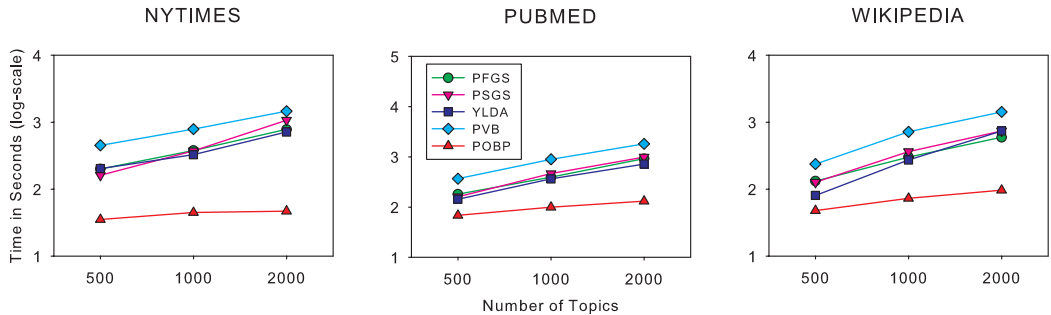}
\caption{
The communication time (in second log-scale) on NYTIMES, PUBMED and WIKIPEDIA using $256$ processors when $K \in \{500, 1000, 2000\}$.
}
\label{communication_ratio}
\end{figure*}

Fig.~\ref{communication_ratio} shows the communication time (in second log-scale) of all algorithms
on NYTIMES, PUBMED and WIKIPEDIA using $256$ processors when $K \in \{500, 1000, 2000\}$.
We see that POBP consumes around $5\% \sim 20\%$ communication time of other algorithms on all data sets.
Among all algorithms,
PVB has the longest communication time because the topic-word distribution $\boldsymbol{\hat{\phi}}_{K \times W}$ in PVB is of single-precision floating type,
leading to an approximately double communication amount than that of GS-based algorithms using integer type.
Although POBP also store $\boldsymbol{\hat{\phi}}_{K \times W}$ in single-precision floating-point format,
it selects only a subset of matrix $\boldsymbol{\hat{\phi}}_{K \times W}$ for communication in subsection~\ref{s3.1}.
Hence,
POBP is more communication-efficient than GS-based algorithms.
According to the analysis in subsection~\ref{s3.2.2},
the total communication time of POBP is proportional to the number of mini-batches $M$.
In our experiments,
the number of mini-batches on NYTIME, PUBMED and WIKIPEDIA is $6$, $19$ and $17$, respectively.
Therefore,
POBP has the least total communication time on NYTIMES.
This result suggests that if the memory is big enough,
we should try to minimize the number of mini-batches $M$ in POBP to reach the minimum communication time.

\subsection{Speed and Scalability}

\begin{figure*}[t]
\centering
\includegraphics[width=0.8\linewidth]{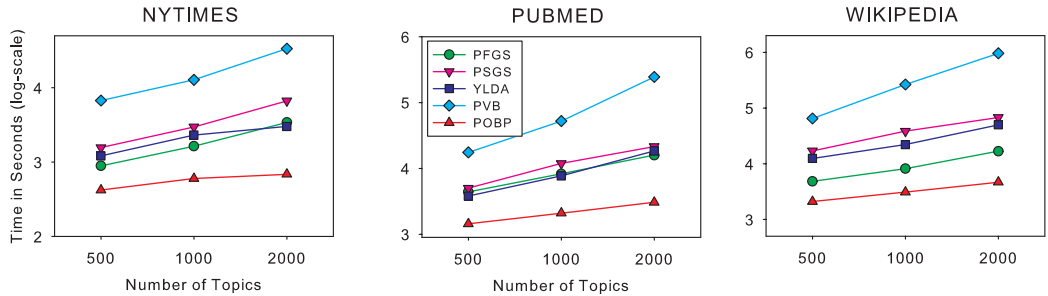}
\caption{
Training time in second (log-scale) of all algorithms on NYTIMES, PUBMED and WIKIPEDIA
when $K \in \{500, 1000, 2000\}$ using $256$ processors.
}
\label{scalability_with_topic}
\end{figure*}

\begin{figure}[t]
\centering
\includegraphics[width=0.8\linewidth]{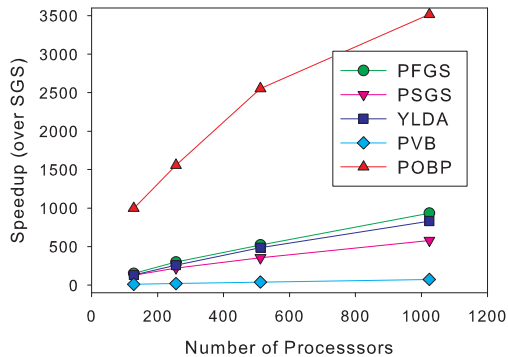}
\caption{
The speedup performance when $K = 2000$.
We choose $1/128$ training time of PSGS~\cite{Yao:09} on $128$ processors as the baseline.
}
\label{k2000_scalability_with_p}
\end{figure}

Fig.~\ref{scalability_with_topic} shows the training time of all algorithms as a function of the number of topics.
We see that POBP is the fastest among all algorithms.
PFGS, PSGS and YLDA have a comparable speed,
and PVB runs the slowest.
On all data sets, POBP is around $5$ to $100$ times faster than other algorithms.
Such a high speed has been largely attributed to three reasons.
First,
POBP has the least communication time as shown in Fig.~\ref{communication_ratio}.
Second,
POBP runs fast at each iteration because it selects the subset of words and topics for computation as shown in Fig.~\ref{pobpcode}.
Finally,
POBP converges very fast as shown in Fig.~\ref{convergence}.

We use the speedup performance with the number of processors~\cite{Newman:09} to evaluate the scalability of parallel algorithms.
We choose the $1/128$ training time of PSGS on $128$ processors as baseline,
which approximates the training time of SGS on a single processor without parallelization.
Then,
the speedup is calculated as the ratio between the baseline and the training time of other parallel algorithms.
Fig.~\ref{k2000_scalability_with_p} shows the speedup performance of all algorithms on PUBMED when $K =2000$.
We show the speedup curve on $N \in \{128, 256, 512, 1024\}$ processors.
Although the speedup curve of POBP bends earlier than other algorithms,
POBP always has much better speedup performance than other parallel algorithms.
This phenomenon confirms that POBP requires only a small number of processors $N^*$ in~\eqref{pobpprocessors} to achieve the best speedup performance,
while other parallel algorithms often need more processors to fulfill it.
Moreover,
the best performance of POBP is much better than those of other algorithms following the analysis in subsection~\ref{s3.2.2}.
In this sense,
POBP has a good scalability because it uses the least number of processors to achieve a much better speedup performance than other parallel algorithms.

\subsection{Memory Usage}

\begin{table}[t]
\centering
\caption{Memory usage (MB) on PUBMED when $K=2000$.}
\begin{tabular}{|c|c|c|c|c|} \hline
$N$                     &PFGS          &PSGS/YLDA       &PVB        &POBP       \\
\hline \hline
$1024$                  &$349$         &$279$           &$438$      &$1,133$  \\
$512$                   &$541$         &$349$           &$560$      &$1,133$ \\
$256$                   &$924$         &$487$           &$804$      &$1,133$ \\
$128$                   &$1,690$       &$765$           &$1,293$    &$1,133$ \\
$64$                    &$N/A$         &$1320$          &$N/A$      &$1,133$ \\
$32$                    &$N/A$         &$N/A$           &$N/A$      &$1,133$ \\
\hline
\end{tabular}
\label{memory}
\end{table}

Big topic modeling tasks are often limited by the memory space of each processor.
Table~\ref{memory} shows the memory usage of all algorithms in each processor on the PUBMED data set when $K=2000$.
The memory usage of PSGS, YLDA, PFGS and PVB decreases with the number of processors,
while POBP consumes a constant memory space.
The major reason is that parallel batch LDA algorithms can distribute both data $\mathbf{x}_{W \times D}$ and
document-topic matrix $\boldsymbol{\hat{\theta}}_{K \times D}$ into $N$ processors,
so that the entire memory usage of each processor will decrease linearly with $N$.
However,
when $N$ is small,
parallel batch LDA algorithms may not load $1/N$ data and document-topic matrix into memory for computation
(e.g., when $N \le 64$, PFGS and PVB fail to process PUBMED in Table~\ref{memory}).
On the other hand,
POBP is an online algorithm that loads only a mini-batch of data and document-topic matrix into memory,
which is a constant value dependent on the mini-batch size $D_m$.
In practice,
users can provide $D_m$ according to each processor's memory quota.
Generally,
we maximize $D_m$ to reduce $M$ for the minimum communication time~\eqref{minicost}.
To further reduce the memory usage of POBP,
we may use hard disk as extended memory to store the word-topic matrix $\boldsymbol{\hat{\phi}}_{K \times W}$ like~\cite{Zeng:14}.
Another strategy is to distribute $\boldsymbol{\hat{\phi}}_{K \times W}$ into $N$ processors by adding more communication costs.
In this way,
we can extract more topics from more vocabulary words without truncation in our experimental settings.

\section{Conclusions} \label{s5}

This paper proposes a novel parallel multi-processor architecture (MPA) for big topic modeling tasks.
This communication-efficient MPA can be combined with both batch and online LDA algorithms.
For example,
we combine this MPA with OBP~\cite{Zeng:14} referred to as the POBP algorithm for big data streams in this paper.
At each iteration,
POBP computes and communicates the subsets of vocabulary words and topics called power words and topics,
and thus has very low computation, memory and communication costs.
Extensive experiments on big data sets confirm that
POBP is faster, lighter, and more accurate than other state-of-the-art parallel LDA algorithms,
such as parallel fast Gibbs sampling (PFGS)~\cite{Porteous:08},
parallel sparse Gibbs sampling (PSGS)~\cite{Yao:09},
Yahoo LDA (YLDA)~\cite{Ahmed:12},
and parallel variational Bayes (PVB)~\cite{Zhai:12}.
Therefore,
POBP is very competitive for web-scale topic modeling applications,
which require a high processing speed under limited resources or seek a high processing efficiency/cost performance.
Since POBP can be interpreted within the EM framework,
its basic idea can be generalized to speed up parallel batch or online EM algorithms for other latent variable models.
Besides,
the power law explanation may shed more light on building faster big learning algorithms
such as deep learning algorithms with high performance computing systems~\cite{Coates:13}.

Future work may include two parts.
First,
we still need to investigate the multi-core architecture (MCA) such as GPU clusters for big topic modeling in the shared memory systems~\cite{Zhuang:13}.
We may avoid serious race conditions by dynamical scheduling of non-conflict subsets of vocabulary words and topics.
Second,
we need to study how to apply POBP in other parallel paradigms like in-memory Map-Reduce (Spark)\footnote{\url{http://spark.incubator.apache.org/}}
or Graph-Lab/Chi.\footnote{\url{http://graphlab.org/}}

\section*{Acknowledgements} \label{s7}

This work is supported by NSFC (Grant No. 61272449, 61202029, 61003154, 61373092 and 61033013),
Natural Science Foundation of the Jiangsu Higher Education
Institutions of China (Grant No. 12KJA520004), Innovative Research
Team in Soochow University (Grant No. SDT2012B02) to JFY and JZ, and a GRF
grant from RGC UGC Hong Kong (GRF Project No. 9041574), a grant from
City University of Hong Kong (Project No. 7008026) to ZQL.

%
%

\bibliographystyle{IEEEtran}
\bibliography{POBP}

\end{document}